\documentclass[letterpaper, 10 pt, conference]{ieeeconf}  

\IEEEoverridecommandlockouts                              

\usepackage{graphicx}

\usepackage[caption=false]{subfig}
\usepackage{xcolor}
\usepackage{array}
\usepackage{float}
\usepackage{booktabs}
\usepackage{adjustbox}
\usepackage[noadjust]{cite}
\usepackage{multirow}
\usepackage{todonotes}
\usepackage{marginnote}

\usepackage[a-1b]{pdfx}

\title{\LARGE \bf
Force Profiling of a Shoulder Bidirectional Fabric-based Pneumatic Actuator for a Pediatric Exosuit}

\author{Mehrnoosh Ayazi,$^{1}$ Ipsita Sahin,$^{2}$ Caio Mucchiani,$^{1}$ Elena Kokkoni,$^{2}$ and Konstantinos Karydis$^{1}$ 
\thanks{$^{1}$~Dept. of Electrical and Computer Engineering; $^{2}$~Dept. of Bioengineering, University of California, Riverside, 900 University Ave, Riverside, CA 92521, USA. Email:{\tt\footnotesize\{mayaz004, isahi001, caiocesr, elenak, karydis\}@ucr.edu}. 
We gratefully acknowledge the support of NSF \# CMMI-2133084. 
Any opinions, findings, and conclusions or recommendations expressed in this material are those of the authors and do not necessarily reflect the views of the National Science Foundation.
}}

\begin{document}
\maketitle
\thispagestyle{empty}
\pagestyle{empty}

\begin{abstract}
This paper presents a comprehensive analysis of the contact force profile of a single-cell bidirectional soft pneumatic actuator, specifically designed to aid in the abduction and adduction of the shoulder for pediatric exosuits. The actuator was embedded in an infant-scale test rig featuring two degrees of freedom: an actuated revolute joint supporting shoulder abduction/adduction and a passive (but lockable) revolute joint supporting elbow flexion/extension. Integrated load cells and an encoder within the rig were used to measure the force applied by the actuator and the shoulder joint angle, respectively. The actuator's performance was evaluated under various anchoring points and elbow joint angles. Experimental results demonstrate that optimal performance, characterized by maximum range of motion and minimal force applied on the torso and upper arm, can be achieved when the actuator is anchored at two-thirds the length of the upper arm, with the elbow joint positioned at a 90-degree angle. The force versus pressure and joint angle graphs reveal nonlinear and hysteresis behaviors. The findings of this study yield insights about optimal anchoring points and elbow angles to minimize exerted forces without reducing the range of motion. 
\end{abstract}

\section{Introduction}
Fabric-based pneumatic actuators~\cite{bhat2023reconfigurable,zhang2023soft} have been integrated into soft robotic devices for assistance and rehabilitation. 
Such devices consider a range of applications including lower limb~\cite{banyarani2024design}, hip~\cite{miller2022wearable}, and upper limb~\cite{xiloyannis2019physiological,missiroli2022rigid, schaffer2024soft,cappello2018assisting} support, grasping~\cite{low2017bidirectional,nguyen2019fabric}, supernumerary robotic limbs~\cite{nguyen2019fabric2}, and haptics~\cite{khin2016soft,zhu2020pneusleeve}. 
Yet, existing efforts have mostly focused on adults~\cite{simpson2017exomuscle, Simpson2020}. 
Only a few have considered developing devices to support upper-extremity (UE) movement in very young children~\cite{Li2019DesignExo, arnold2020exploring, Kokkoni2020_asme}. 


To ensure comfortable and safe interaction between the user and the device, it is crucial to understand the exerted forces.   
Exerted forces can be determined based on the device's mechanical components and affordances, such as actuator deformation as a function of input pressure~\cite{simpson2017exomuscle} and anchoring~\cite{wei2018design}. 
Desired values for these forces are determined based on the users' needs and tolerances~\cite{ramirez2019jacket}. 
Previous studies have focused on determining desired values to support motion about different joints and then developing adult devices capable of generating such forces.  
For example, the average peak torque requirement produced by a healthy deltoid muscle to completely elevate the arm was found to be $43$\;Nm~\cite{de2008torque}. 
To reach maximum arm elevation, a force of at least $270$\;N must be generated~\cite{arellano2019soft}. 
Based on this analysis, a soft wearable deltoid assistance device~\cite{arellano2019soft} was developed to produce about $300$\;N force on the upper arm to elevate the arm to $90^{\circ}$. 
An exomuscle~\cite{simpson2017exomuscle} can exert forces on the arm in the range of $[0-500]$\;N with varying actuator inflating pressure and shoulder abduction angle while testing on a one-degree-freedom apparatus. 
Other studies like a soft shoulder assistive device with pneumatic artificial muscles~\cite{sridar2018soft} and a cable-driven soft orthotic device~\cite{kesner2011design} have targeted the generation of a maximum force output of approximately $100$\;N, which is similar to the compressive force generated by biological anterior deltoid muscle. 
Despite these findings, there have been no studies so far on the forces exerted by actuators or soft wearable UE devices on the infant body. 


\begin{figure}[!t]
\vspace{6pt}
     \centering
     \includegraphics[trim={0cm 0cm 0cm 0.5cm},clip,width=0.80\columnwidth]{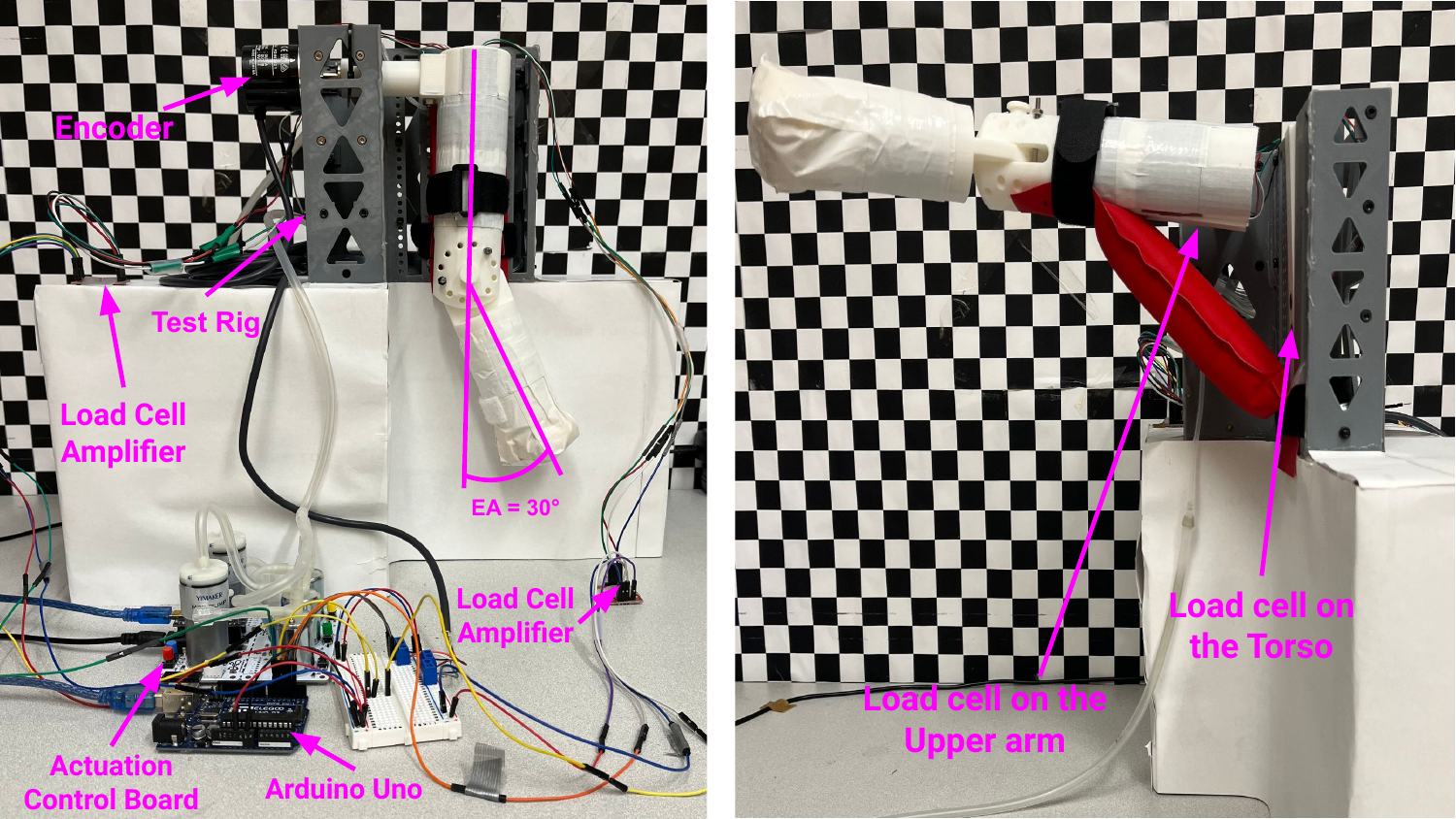}
     \vspace{-6pt}
         \caption{Experimental setup employed in this work for force profiling of a fabric-based pneumatic shoulder actuator when the actuator anchoring points and elbow (locked in place) angle vary.}
     \label{fig:setup}
     \vspace{-21pt}
\end{figure}

Our prior work has focused on the design and kinematics of actuators to support UE movement in very young children~\cite{sahin2022bidirectional, sahin2023fabric}, as well as on the development of feedback control methods for exosuit prototypes employing those actuators tested with an engineered mannequin~\cite{mucchiani2022closed,mucchiani2023robust}.
In this work, we study the force generation of our previously developed fabric-based pneumatic shoulder actuator using a static testing rig (Fig.~\ref{fig:setup}). 
We vary two key conditions---anchoring points of the actuator on the torso and upper arm, and elbow joint angle---and observe how the exerted force and range of motion are affected. 
The goal is to identify which conditions can lead to minimal exerted forces and maximal range of motion. 
This information, in turn, can help pinpoint how to improve the exosuit's functionality.

\section{Materials and Methods}

\subsection{Actuator Description}
The soft pneumatic actuator used in this study was a rectangular-shaped, single-cell actuator, crafted from flexible thermoplastic polyurethane (TPU) fabric (Oxford 200D heat-sealable coated fabric with a thickness of $0.020$\;cm). 
The inflatable portion of the actuator is $15\times5$\;cm. 
Two non-inflatable segments of about $3\times5$\;cm at each end are used to facilitate attachment to the torso and upper arm using velcro straps. 
This actuator can support shoulder abduction and adduction without impeding other degrees of freedom (DoFs) at the joint by being placed in a low-profile position within the axilla. 
Among various actuators evaluated for shoulder assistance~\cite{sahin2022bidirectional}, this variant was selected based on rigorous criteria, with a primary focus on range of motion (ROM).

\subsection{Hardware Experimentation Setup}
We developed a 3D-printed test rig modeled based on the UE of the 50th percentile of a 12-month-old infant~\cite{edmond2020normal,Fryar2021} (Fig.~\ref{fig:setup}). 
The rig comprises a base (torso), an upper arm, a forearm, and a weighted attachment ($0.06$\;kg) at the end of the forearm to emulate the hand. 
The upper arm is connected to the base via a revolute joint acting as the shoulder joint.  
The forearm is connected to the upper arm through another revolute joint, acting as the elbow joint, which can be locked in place at four distinct angles ($EA\in\{0^{\circ},30^{\circ},60^{\circ},90^{\circ}\}$). 
The upper arm measures $16.4$\;cm in length and $14.7$\;cm in circumference, while the forearm measures $10.85$\;cm in length and $14.51$\;cm in circumference. 
Both the upper arm and forearm are hollow, and sandbags were embedded within to match the weight of an infant's upper arm and forearm (total weight about $0.432$\;kg), based on the infant anthropometrics for a total body mass of $10.8$\;kg~\cite{zernicke1992mass}.

Actuator inflation/deflation was controlled by an external pneumatic board (Programmable-Air kit). 
The board has two compressor/vacuum pumps and three pneumatic valves to regulate airflow ($2$\;L/min). 
Airflow is adjusted via the pump duty cycle ($[0-100]$\%). 
The pressure range is $[-50, 50]$\;kPa. 
The board also incorporates a pressure sensor (SMPP-03) and an Arduino Nano (ATMega328P) for control and monitoring. 
The embedded Arduino facilitates connectivity to a computer through a serial-to-USB interface, allowing users to send commands to the pumps and valves, and log sensor data.

The developed force measurement system consisted of two $5$\;kg load cells (YZC-133) embedded in the upper arm and base of the test rig. 
The upper arm load cell was positioned along the vertical centerline on the inner side of the upper arm. 
A plate measuring $11.1\times 1.36$\;cm was attached to the load cell to measure the force applied to the area covered by the plate. 
The load cell on the base was aligned with the vertical centerline on the side facing the inner side of the upper arm and was coupled with a $15.6\times 1.66$\;cm plate. 
Each load cell was connected to an amplifier (HX711, Sparkfun); data were read at a frequency of $10$\;Hz. 
A 1024 pulse per rotation rotary encoder (E6B2, Sparkfun) was used to measure the angle of the shoulder joint throughout the experiments. 
The amplified outputs of load cells and the encoder's data were transmitted to a computer through another microcontroller (Arduino UNO). 

\subsection{Experimental Conditions}
The performance of the actuator in terms of force exerted on the upper arm and torso was assessed within a total of eight different conditions varying based on the upper arm anchoring point and the locked elbow angle (Table~\ref{tab:conditions}). 
A total of 30 trials were conducted for each condition. 
The actuation control board was operating at a $100\%$ duty cycle to achieve the fastest possible inflation and deflation. 
A complete reaching motion in infants lasts about two seconds~\cite{zhou2021infant}; hence, we assessed both inflation and deflation phases each lasting for $5$ seconds, starting with inflation. 
Internal actuator pressure peaked at a maximum of $34$\;kPa and dropped to a minimum of $-34$kPa when fully inflated and deflated, respectively.

\begin{table}[htb]
\vspace{-6pt}
\caption{Experimental Conditions}\label{tab:conditions}
\vspace{-12pt}
\begin{center}
\renewcommand{\arraystretch}{1.2} 
\begin{tabular}{c c}
\toprule
Anchoring Point & Elbow Angle\\
\midrule
{$\frac{2}{3}$ length of UA} & 0$^{\circ}$,  30$^{\circ}$ , 60$^{\circ}$, 90$^{\circ}$ \\
\midrule
{$\frac{1}{2}$ length of UA} & 0$^{\circ}$,  30$^{\circ}$,  60$^{\circ}$,  90$^{\circ}$ \\
\bottomrule
\end{tabular}
\end{center}
\vspace{-18pt}
\label{table:one}
\end{table}

\section{Results and Discussion}


\begin{figure}[!t]
\vspace{6pt}
     \centering
     \includegraphics[width=1\columnwidth]{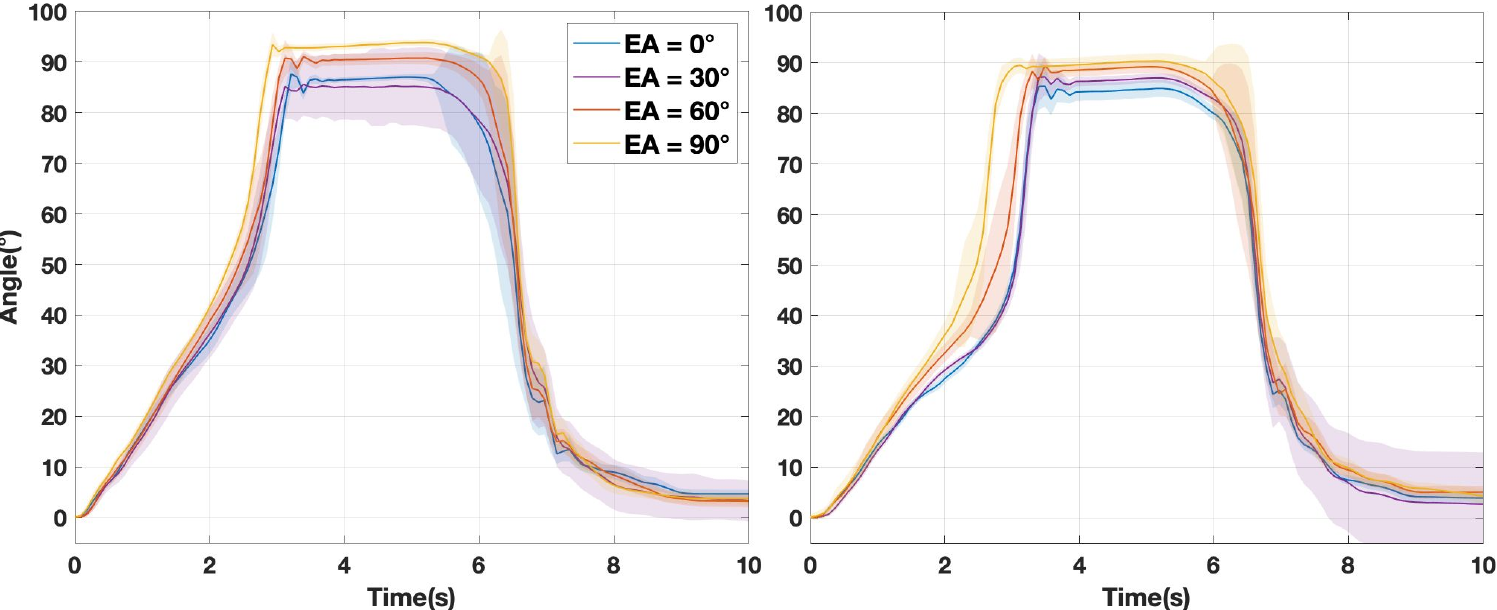 }
     \vspace{-20pt}
         \caption{Changes in shoulder joint angle over time where shaded area represents standard deviation. The panels at left and right represent anchoring at $\frac{2}{3}$ and $\frac{1}{2}$ the length of the upper arm, respectively.}
     \label{fig:angle vs. time}
     \vspace{-21pt}
\end{figure}

\subsection{Assessment of Shoulder Joint ROM}
We first assessed shoulder joint ROM under the different conditions listed in Table~\ref{tab:conditions}. 
It can be observed (Fig.~\ref{fig:angle vs. time}) that increasing the elbow joint angle (i.e. flexing the elbow) results in a larger maximum shoulder joint angle as well as increased ROM. 
During elbow flexion, there is a reduction of the moment arm between the point where the mass at the end of the forearm is located and the shoulder joint axis of rotation. 
Less mechanical work is thus required but the amount of input pressure remains the same, which leads to larger ROM. 
Further, the shoulder joint has larger ROM when the actuator is anchored at $\frac{2}{3}$ the UA length. 
The moment arm between the actuator force location on the upper arm and shoulder joint increases when the placement of the actuator moves from half to two-thirds of UA length. Consequently, the actuator generates more torque with the same actuator pressure and force on the upper arm.

\subsection{Assessment of Actuator Force Generation}
\begin{figure}[!t]
\vspace{2pt}
     \centering
     \includegraphics[width=1\columnwidth]{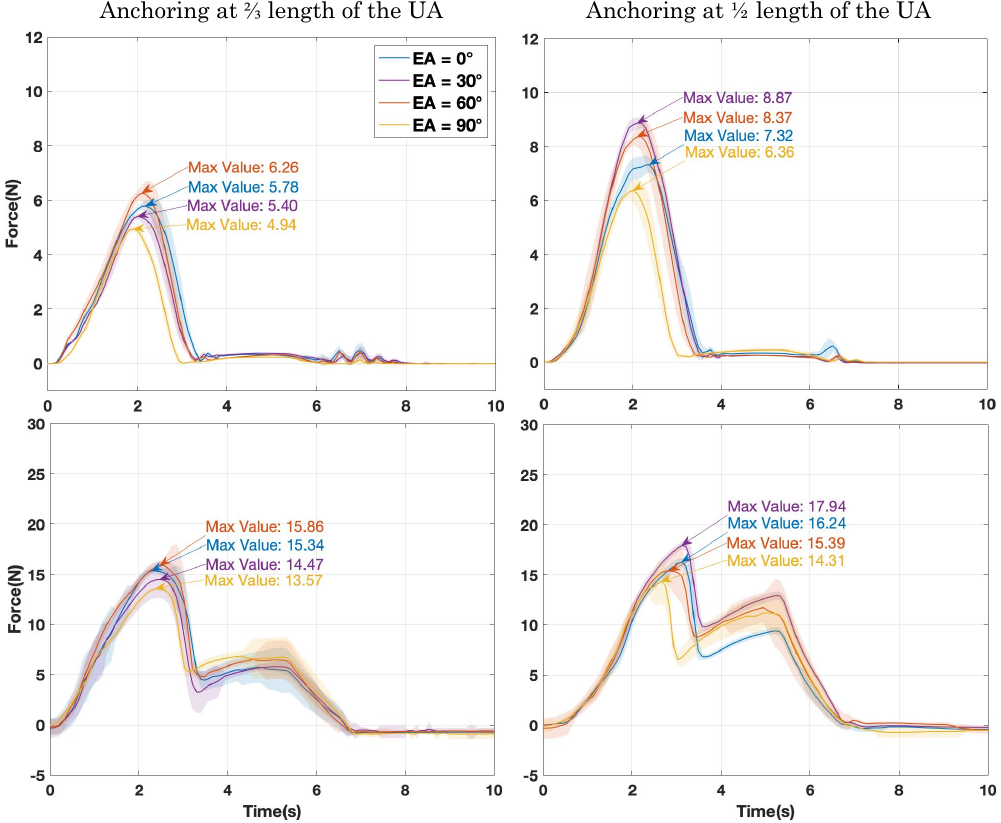}
     \vspace{-20pt}
         \caption{Forces applied by the actuator on the torso (Top) and upper arm (UA) (Bottom) for the different experimental conditions listed in Table~\ref{tab:conditions}. Shaded zones reflect one standard deviation (30 trials for each condition).
         }
     \label{fig:force vs. time}
     \vspace{-21pt}
\end{figure}

We then examined the variation in the force exerted by the actuator on the torso and upper arm. 
Generally, the maximum (on average) force applied to the torso and upper arm decreases as the elbow joint angle increases (Fig.~\ref{fig:force vs. time}). 
However, this trend is not observed across all cases. 
The peak force on the torso and upper arm is created by the interaction between two segments of the actuator, which apply force to these areas. 
As the elbow flexes and the moment of inertia decreases, the interaction force between the segments decreases, thus reducing the maximum force applied to both the upper arm and torso. 
Further, when comparing peak forces between the two anchoring points, a notable trend emerges: while the force on the torso increases, the force on the upper arm stays relatively consistent. 
As the anchoring point shifts from two-thirds to half the length of UA, the area covered by the actuator on the torso expands which results in a larger force applied to the torso with the same actuator inner pressure. 
Since the actuator maintains the same coverage area on the upper arm in both scenarios, the maximum force exerted on the upper arm does not change.

The decline in force on both the upper arm and torso during actuator inflation can be associated with varying contacts during the process. 
Placed under the armpit, a single-cell actuator naturally folds, creating two distinct sections that effectively function as two separate cells until a critical point when they merge. 
These sections interact by pushing outward while inflating, increasing force on both the torso and upper arm.
Around halfway through inflation, these sections merge into a single cell (crucial for maximizing ROM). 
However, this merging leads to a notable decrease in contact with both the upper arm and torso. Consequently, force at both points decreases abruptly. 
After unfolding, the actuator remains attached to the torso at its non-inflatable end, causing the force measured at the torso to drop nearly to zero. 
The force on the upper arm also declines but at a smaller degree since the inflatable part of the actuator still partially covers the upper arm. 
Hence, the force at the upper arm increases until the end of the inflation period.
During deflation, the upper arm force decreases while the torso force remains near zero.

\begin{figure}[!t]
\vspace{2pt}
     \centering
     \includegraphics[width=1\columnwidth]{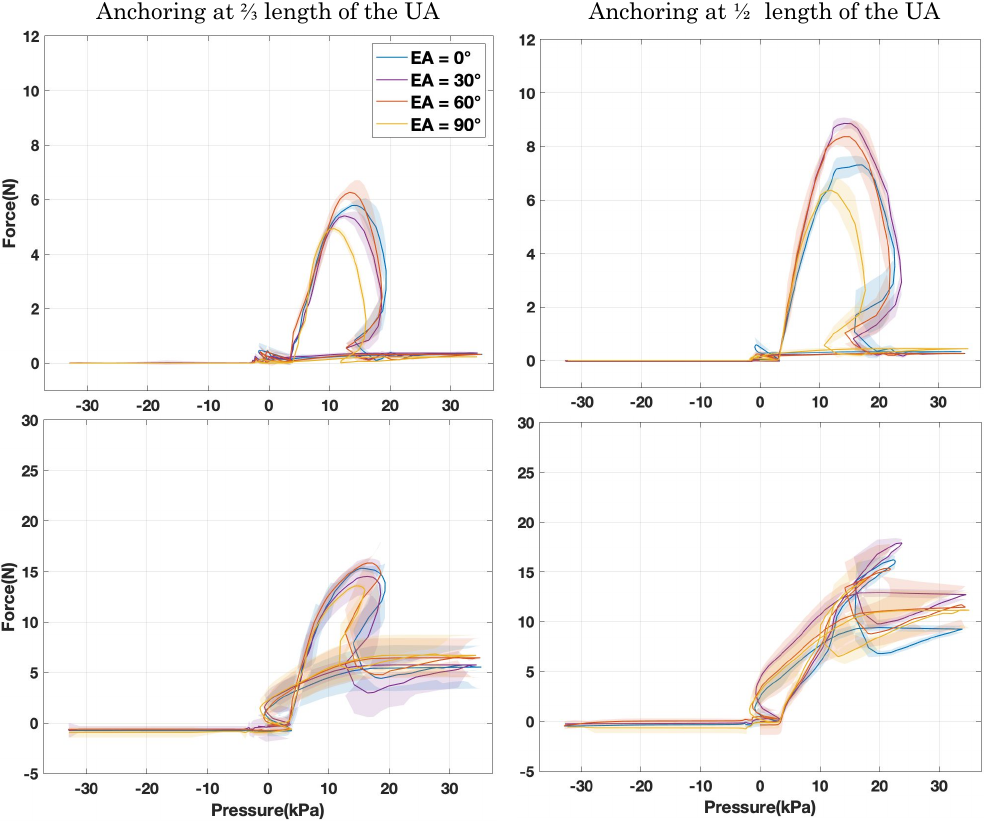}
     \vspace{-20pt}
         \caption{Actuator force changes on the torso (Top) and upper arm (Bottom) as the internal pressure of the actuator varies. 
         }
     \label{fig:force vs. pressure}
     \vspace{-9pt}
\end{figure}

\begin{figure}[!t]
\vspace{0pt}
     \centering
     \includegraphics[width=1\columnwidth]{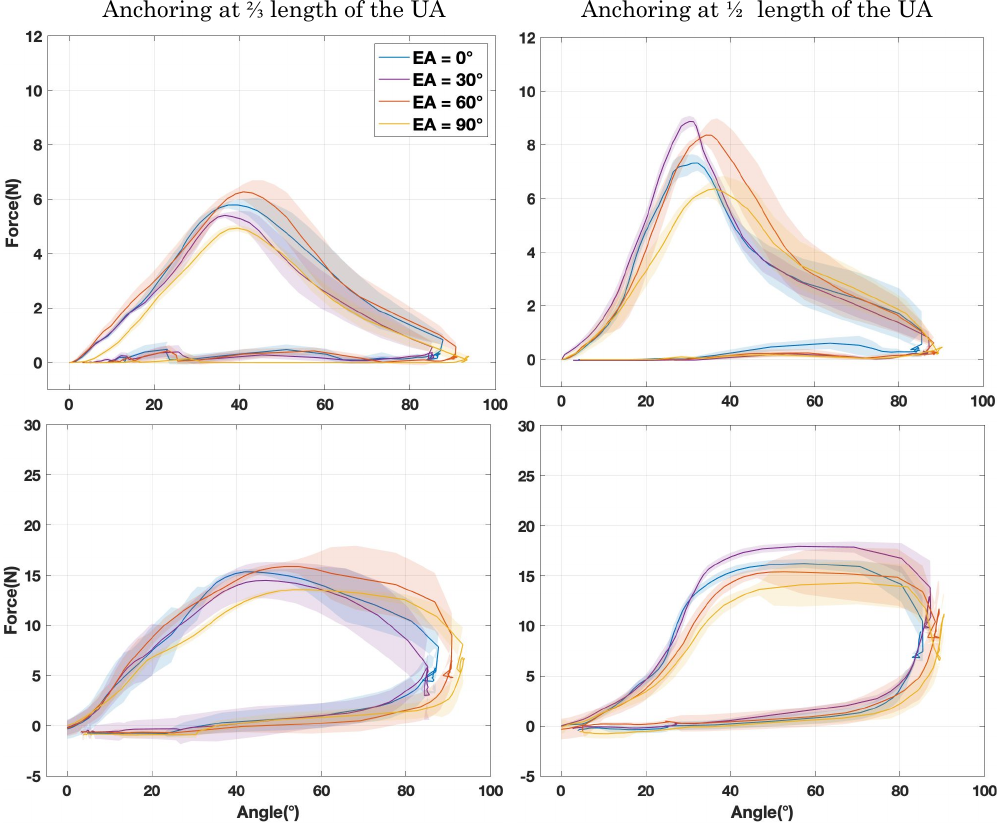}
     \vspace{-20pt}
         \caption{Actuator force changes on the torso (Top) and upper arm (Bottom) as the shoulder joint angle increases. 
         }
     \label{fig:force vs. angle}
     \vspace{-21pt}
\end{figure}


We also analyzed the relation between the forces exerted on the upper arm and torso and the actuator's internal pressure.
The forces were found to be nonlinear and exhibit hysteresis as the input pressure varied (Fig.~\ref{fig:force vs. pressure}). 
For pressures below $10$\;kPa, the relation is linear; however, around $10$\;kPa, the actuator begins to unfold and a notable drop occurred because of the reduction in the contact area. 
The actuator pressure also decreases, likely due to the suddenly increased volume from the actuator unfolding. 
After this point, the force on the torso remains unchanged while the force on the upper arm starts to increase linearly with pressure. 
During deflation, the force on the torso is constant, whereas the force on the upper arm decreases roughly logarithmically.

Further, we analyzed the relation between exerted force and shoulder joint angle (Fig.~\ref{fig:force vs. angle}). 
There is a major change when at about half the ROM, marking the point where the actuator begins to unfold. 
Identifying this angle as the threshold for actuator unfolding, we found that the force behavior aligns with the patterns described previously. 
Nonlinearity and hysteresis were observed here as well.

The maximum force generated by the actuator in this work is significantly lower compared to those developed for adult shoulder abduction/adduction~\cite{simpson2017exomuscle,arellano2019soft} which are at the range of $[300-500]$\;N. 
Although a direct comparison of absolute values may not be appropriate, this information may help suggest upper bounds for exerted forces via a mechanics-based scaling. 
Such an assessment merits further investigation considering that there is limited information in the literature regarding the maximum forces that are safe to exert on infants' arms in the context of pediatric exosuits.

\section{Conclusion}
This study examined the contact forces generated by a single-cell bidirectional soft pneumatic actuator for shoulder abduction/adduction using an infant-scale test rig. 
Extensive experimentation across different conditions related to actuator anchoring and (fixed) elbow joint angle demonstrated the optimal performance range of the considered actuator: the maximum range of motion while exerting minimal force on the torso and upper arm is achieved when anchored at two-thirds the length of the upper arm when the elbow joint is at a $90^{\circ}$ angle. 
Nonlinearity and hysteresis between actuator input pressure, shoulder joint angle, and the exerted forces were identified. 
These were caused by the actuator's rapid unfolding during actuation, thereby suggesting future design improvements to produce smoother force profiles.

\bibliography{ipsita, Kokkoni, caio}
\bibliographystyle{IEEEtran}

\end{document}